%% file: main.tex
\documentclass[10pt, conference]{IEEEtran}
\IEEEoverridecommandlockouts
\usepackage{cite}
\usepackage{amsmath,amssymb,amsfonts}
\usepackage{graphicx}
\usepackage{textcomp}
\usepackage{xcolor}
\usepackage{bm}
\usepackage{dsfont}
\usepackage{comment}
\usepackage{algorithm}
\usepackage{algpseudocode}
\usepackage{tikz}
\usepackage[nice]{nicefrac}
\usepackage{multirow}

\newcommand\copyrighttext{%
  \footnotesize \textcopyright 2023 IEEE. Personal use of this material is permitted.  Permission from IEEE must be obtained for all other uses, in any current or future media, including reprinting/republishing this material for advertising or promotional purposes, creating new collective works, for resale or redistribution to servers or lists, or reuse of any copyrighted component of this work in other works.
 
  Accepted at 2023 ACM/IEEE International Symposium on Low Power Electronics and Design (ISLPED).}
\newcommand{\copyrightnotice}{%
\begin{tikzpicture}[remember picture,overlay,scale=1.00, every node/.style={scale=1.00}]
\node[anchor=south,yshift=10pt] at (current page.south) {\fbox{\parbox{\dimexpr\textwidth-\fboxsep-\fboxrule\relax}{\copyrighttext}}};
\end{tikzpicture}%
}

\definecolor{LightGray}{gray}{0.9}
\definecolor{darkspringgreen}{rgb}{0.09, 0.45, 0.27}
\def\BibTeX{{\rm B\kern-.05em{\sc i\kern-.025em b}\kern-.08em
    T\kern-.1667em\lower.7ex\hbox{E}\kern-.125emX}}

\begin{document}
\bstctlcite{IEEEexample:BSTcontrol}
\title{Precision-aware Latency and Energy Balancing on Multi-Accelerator Platforms for DNN Inference
}

\author{\IEEEauthorblockN{Matteo Risso\IEEEauthorrefmark{1}, Alessio Burrello\IEEEauthorrefmark{1}\IEEEauthorrefmark{2}, Giuseppe Maria Sarda\IEEEauthorrefmark{3}, Luca Benini\IEEEauthorrefmark{2}, \\ Enrico Macii\IEEEauthorrefmark{1}, Massimo Poncino\IEEEauthorrefmark{1}, Marian Verhelst\IEEEauthorrefmark{3}, Daniele Jahier Pagliari\IEEEauthorrefmark{1}}
\IEEEauthorblockA{\IEEEauthorrefmark{1}Politecnico di Torino, Turin, Italy. \IEEEauthorrefmark{2}University of Bologna, Bologna, Italy. \IEEEauthorrefmark{3}KU Leuven, Belgium.
}\textit{Corresponding Email: matteo.risso@polito.it}
}

\maketitle

\copyrightnotice

\input{text/Abstract}

\input{text/Intro}
\input{text/Background}
\input{text/Methods}
\input{text/Results}
\input{text/Conclusion}

\footnotesize
\bibliographystyle{IEEEtran}

\end{document}

%% file: text/Abstract.tex
\begin{abstract}
The need to execute Deep Neural Networks (DNNs) at low latency and low power at the edge has spurred the development of new heterogeneous Systems-on-Chips (SoCs) encapsulating a diverse set of hardware accelerators. 
How to optimally map a DNN onto
such 
multi-accelerator
systems is an open problem.
We propose ODiMO, a hardware-aware tool that performs a fine-grain mapping across different accelerators on-chip, splitting individual layers and executing them in parallel, to reduce inference energy consumption or latency, while taking into account each accelerator's quantization precision to maintain accuracy.
Pareto-optimal networks in the accuracy vs. energy or latency space are pursued for three popular dataset/DNN pairs, and deployed on the DIANA heterogeneous ultra-low power edge AI SoC. We show that ODiMO reduces energy/latency by up to 33\%/31\% with limited accuracy drop (-0.53\%/-0.32\%)
compared to manual heuristic mappings.
\end{abstract}

%% file: text/Intro.tex
\section{Introduction}\label{sec:intro}
Executing Deep Neural Networks (DNNs) inference at the edge 
brings several advantages, including lower energy consumption, lower and more predictable response latency, and improved privacy, by eliminating the dependency on a constant Internet connection~\cite{Zhou2019, sze2020efficient}. However, deploying computationally intensive DNNs on edge devices with tight power envelopes and energy constraints is a daunting task, addressed by current research in two orthogonal ways. On the software side, optimization techniques such as constrained Neural Architecture Search (NAS), pruning, and quantization~\cite{sze2020efficient,Jacob2018}, are applied to DNN models to make them both accurate and resource-efficient. On the hardware side, efficiency is improved through specialization, i.e., by designing increasingly heterogeneous Systems-on-Chip (SoCs), equipped with domain specific accelerators for DNN processing~\cite{edgetpu,Dagli2022,hp_ssc2020, ueyoshi2022diana}.
In particular, a recent trend goes towards \textit{multi-accelerator} SoCs,
in which multiple specialized hardware blocks are
either optimized for different DNN operations, or to perform the same operations with different trade-offs in terms of latency, throughput, energy efficiency or accuracy~\cite{hp_ssc2020,ueyoshi2022diana,Dagli2022}. 

How to optimize a DNN model for execution onto these multi-accelerator systems is an open problem. In fact, classic model optimizations
are either hardware-independent targeting abstract complexity metrics, or tailored to the scenario in which the entire network runs on a single device (CPU, GPU, etc).
While more recent works considered multi-device inference~\cite{Wang2020,Vasiliadis2022,Tu2019,Dagli2022,Jeong2022,Song2020,Kang2017,JahierPagliari2020b}, to our knowledge, they all assumed that all devices could produce equivalently accurate results. This is not always true, with a key counter-example being SoCs including both Digital and Analog In-Memory-Computing (AIMC) accelerators~\cite{hp_ssc2020,ueyoshi2022diana}, where the latter can be faster and more efficient, but produce approximated results due to very low quantization bit-width used for weights (e.g., binary or ternary), while the former are slower and more energy hungry, but process wider data items at higher numerical precision.

In this work, we propose a novel approach to optimize and map a DNN execution onto such kind of system, which takes into account the quantization supported by different accelerators already at training time. Namely, we leverage a fine-grained, gradient-based, mixed-precision search method~\cite{cai2020rethinking,risso2022channel} to partition each DNN layer onto sub-layers, executed in parallel by the various accelerators using their respective precision. While taking into account the possible accuracy drops due to quantization, our method tries to 
minimize energy consumption or latency, through appropriate hardware-aware cost models.
We name our approach \textbf{O}ne-shot \textbf{Di}fferentiable \textbf{M}apping \textbf{O}ptimizer (\textbf{ODiMO}).

\looseness=-1
With experiments on three popular Convolutional Neural Network (CNN) architectures, trained on edge-relevant computer vision benchmarks, we show that our method yields rich Pareto-fronts of mapping solutions in the accuracy versus latency or energy spaces, under different assumptions regarding the accelerators capabilities in a heterogeneous SoC. When deployed on a real-world SoC of this kind, DIANA~\cite{ueyoshi2022diana}, our optimized models reduce energy/latency by up to $33\%/31\%$ with limited accuracy drops (-$0.53\%/$-$0.32\%$) compared to manual mappings based on rules of thumb. Furthermore, we improve accuracy by up to +$37\%$ for a 1.12$\times$ energy increase compared to a solution that only tries to minimize energy without considering accuracy.
Our code is open-sourced at: \texttt{https://github.com/eml-eda/odimo}.

%% file: text/Background.tex
\section{Background and Related Works}
\label{sec:background}
\subsection{Specialized hardware for edge DNN inference}\label{sec:soc}
In recent years, specialized architectures for DNN processing at the edge have flourished, with several designs proposed both in industry and academia~\cite{reutherAI2021}.
Many of these modern SoCs contain \textit{multiple} specialized hardware blocks, able to execute the same workload with different trade-offs in terms of latency, throughput, energy consumption, or accuracy.
One example is the Jetson AGX Xavier series from NVIDIA, equipped with an 8-cores ARM CPU, a NVIDIA Volta GPU with 512 CUDA cores and two NVIDIA Deep Learning Accelerators (NVDLA). Users can split the workload between the GPU, faster but more energy hungry, and the NVDLAs, slightly slower but more efficient~\cite{Dagli2022}. 

In the architecture of \cite{hp_ssc2020}, a control CPU dispatches the workload either to a 590k-cells AIMC accelerator tailored for 1-bit multiply-and-accumulate (MAC) operations, or to a digital Near-Memory Computing (NMC) accelerator, which supports variable precision from 1 to 8bits.
In this case, selecting one of the two accelerators results either in more accuracy but higher latency and energy (NMC), or vice versa (AIMC).
Similarly, DIANA~\cite{ueyoshi2022diana} features a single-core RISC-V CPU as control unit and two DNN accelerators: a 16$\times$16 grid of digital processing elements performing MACs at 8-bit precision, with a 64 kB weight memory, and a 500k-cells AIMC accelerator with ternary weights. The two accelerators share a dedicated 256 kB L1 memory, accessed through Direct Memory Access (DMA).

\subsection{Mixed-Precision Quantization} %
In parallel to new specialized SoCs, many DNN optimization techniques have been introduced over the years, such as pruning, quantization, and NAS, to design lightweight networks that can fit on edge devices.
This section focuses on the main knob explored by ODiMO, i.e., quantization; we refer readers to~\cite{sze2020efficient} for details on other techniques.

Quantization improves DNNs' energy-efficiency by reducing the precision of data and operations, e.g., from floating point to low bit-width integer formats (1 to 8-bit)~\cite{Jacob2018}.
The default approach is the so-called \textit{fixed-precision} quantization, in which the same bit-width $n$ (usually 8-bit) is used across the model. 
Recently, \textit{mixed-precision} approaches, that vary $n$ for different parts of a DNN, have been shown to provide additional time, memory and energy savings, especially when native hardware support for sub-byte operations is available~\cite{wang2019haq,dong2020hawq,cai2020rethinking,risso2022channel}. However, finding the optimal assignment of bit-widths to different parts of the network, e.g., to minimize energy under a given accuracy constraint, involves searching a huge space, exponential in the depth of the DNN.

Existing solutions to this problem use techniques inherited from NAS, such as sensitivity-based heuristics~\cite{dong2020hawq} or Reinforcement Learning~\cite{wang2019haq}. In particular, a recent approach~\cite{cai2020rethinking,risso2022channel} takes inspiration by Differentiable NAS (DNAS) to speed up the process, optimizing the bit-width assignment \textit{during training}.
Multiple copies of each tensor, quantized at different bit-widths, are generated on-the-fly, and linearly combined by means of trainable NAS parameters. The latter are then inserted in a standard DNN training loop, where an appropriately regularized loss function guides the optimization to increase the NAS parameters linked with quantizations that yield a good trade-off between accuracy and inference cost.
At the end of training, the bit-widths which have been assigned the largest NAS coefficient are selected for each tensor.
\subsection{DNN mapping on heterogeneous systems}\label{sec:related}
The problem of mapping complex tasks onto a heterogeneous system with accelerators has been studied for a long time. Early works focus on generic workloads (e.g. OpenCL programs)~\cite{Konrad2018}, but more recently, the specific case of DNN inference has attracted a lot of attention. The authors of \cite{Wang2020} implement a simple form of data parallelism, in which entire inferences are mapped onto a single device, selecting the fastest available between CPUs, GPUs and NPUs at any time. \cite{Vasiliadis2022} performs a similar mutually-exclusive mapping, but at the level of each DNN layer, using a random forest to predict the latency or energy efficiency of offloading a layer to CPU or to multiple GPUs, based on the tensors geometry.
\cite{Tu2019} proposes a heuristic for a multi-accelerator system including a GPU (NVIDIA Jetson TX2) and a FPGA (Xilinx Artix7), consisting in offloading all Fully-Connected (FC) layers to the FPGA, and the rest of the DNN to the GPU. 

The authors of \cite{Dagli2022} explore the energy versus latency trade-offs offered by offloading parts of a DNN to the GPU or to the NVDLAs in a NVIDIA Jetson AGX Xavier. Partitioning is done at layer level, and linear programming is used to find the lowest latency mapping under an energy constraint. An alternative mapping scheme for the Xavier is proposed in \cite{Jeong2022}, which explores data parallelism and pipelining among GPU and NVDLAs, focusing only on improving throughput.

In \cite{Song2020}, finer-granularity intra-layer partitions are explored, using dynamic programming to optimize DNN training latency on a system composed of multiple Google TPUv2/v3 accelerators, taking into account compute performance and communication overheads. Three partitioning axes are considered (over batches, input channels or output channels). Lastly, other works target DNN mapping problems for networks of distributed devices rather than individual multi-accelerator SoCs, proposing similar data- or model-partitioning schemes~\cite{Kang2017,JahierPagliari2020b}.

%% file: text/Methods.tex
\section{One-shot Differentiable Mapping Optimizer}
All works discussed in see Sec.~\ref{sec:related} assume that mapping part of a DNN to a given accelerator does not affect the final accuracy~\cite{Vasiliadis2022,Wang2020,Tu2019,Dagli2022,Jeong2022,Song2020,Kang2017,JahierPagliari2020b}. 
Therefore, they only explore the trade-off between latency and throughput or latency and energy.
While reasonable for their targets, this assumption breaks for more extreme-edge-oriented platforms such as~\cite{hp_ssc2020,ueyoshi2022diana} in which some of the available accelerators use aggressive quantization. 
In that case, the mapping choices greatly influence both functional (accuracy) and non functional (e.g., energy) metrics.
To our knowledge, no previous work has considered this trade-off in DNN mapping optimizations for multi-accelerator SoCs.

We fill this gap by proposing ODiMO, an optimization method
that partitions a DNN execution onto heterogeneous compute domains that include accelerators with different quantization levels and formats, optimizing the trade-off between accuracy and energy or latency.
Differently from most conventional mapping strategies, which are coarse-grain (e.g., layer-wise), our tool uses a fine-grain intra-layer partitioning aimed at maximizing the utilization of all accelerators.
\subsection{Mapping optimization strategy}\label{subsec:strategy}
\begin{figure}[t]
  \centering
  \includegraphics[width=.9\columnwidth]{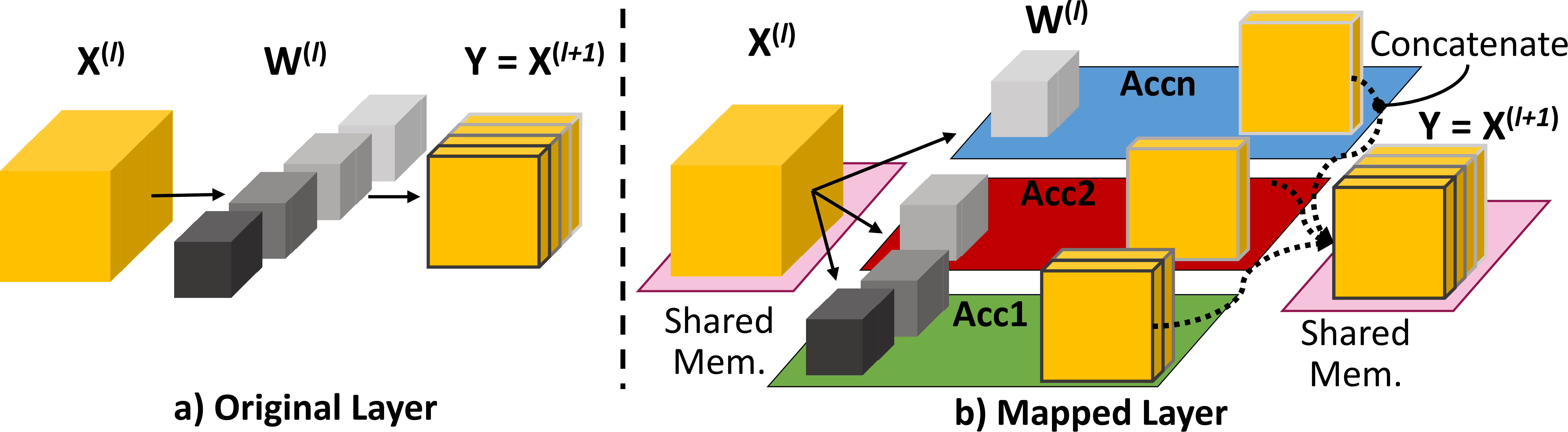}
  \vspace{-0.2cm}
  \caption{Mapping strategy for a Convolutional layer.}
  \label{fig:map}
    \vspace{-0.6cm}
\end{figure}

ODiMO considers splitting each Convolutional (Conv) or FC layer in a DNN among $N$ different devices, at the level of \textit{individual output channels/neurons}. That is, all  accelerators process the entire layer input and produce a subset of the output activations, as shown in Figure \ref{fig:map} for a Conv layer.
For this approach to be effective, the target heterogeneous system must respect two properties: i) different accelerators can have incompatible \textit{weights} quantizations but must have the same \textit{activation} quantization, or at least two similar formats that do not cause a significant difference in terms of accuracy (see Sec.~\ref{subsec:training}); ii) all accelerators must have access to a shared memory for loading/storing the layer input/partial output~\cite{Song2020}. Note that both~\cite{hp_ssc2020} and \cite{ueyoshi2022diana} respect these constraints.
Under these conditions, the problem can be reduced to selecting the best quantization for each channel's weights, where the choice not only influences the overall model accuracy, but also limits the mapping options for that channel to the accelerator(s) that support the selected precision, thus affecting the inference energy/latency costs.

The optimization space is huge: e.g., for just $N=2$ accelerators and a ResNet18 CNN, there are about $10^{39}$ possible ways to assign each channel of each layer to one of the two devices.
Therefore, ODiMO adopts a DNAS-like optimization method inspired by recent work on fine-grained mixed-precision quantization~\cite{risso2022channel}, in which bit-width assignment is performed \textit{during training}, similar to so-called One-shot NAS approaches.

\begin{figure}[t]
  \centering
  \includegraphics[width=.95\columnwidth]{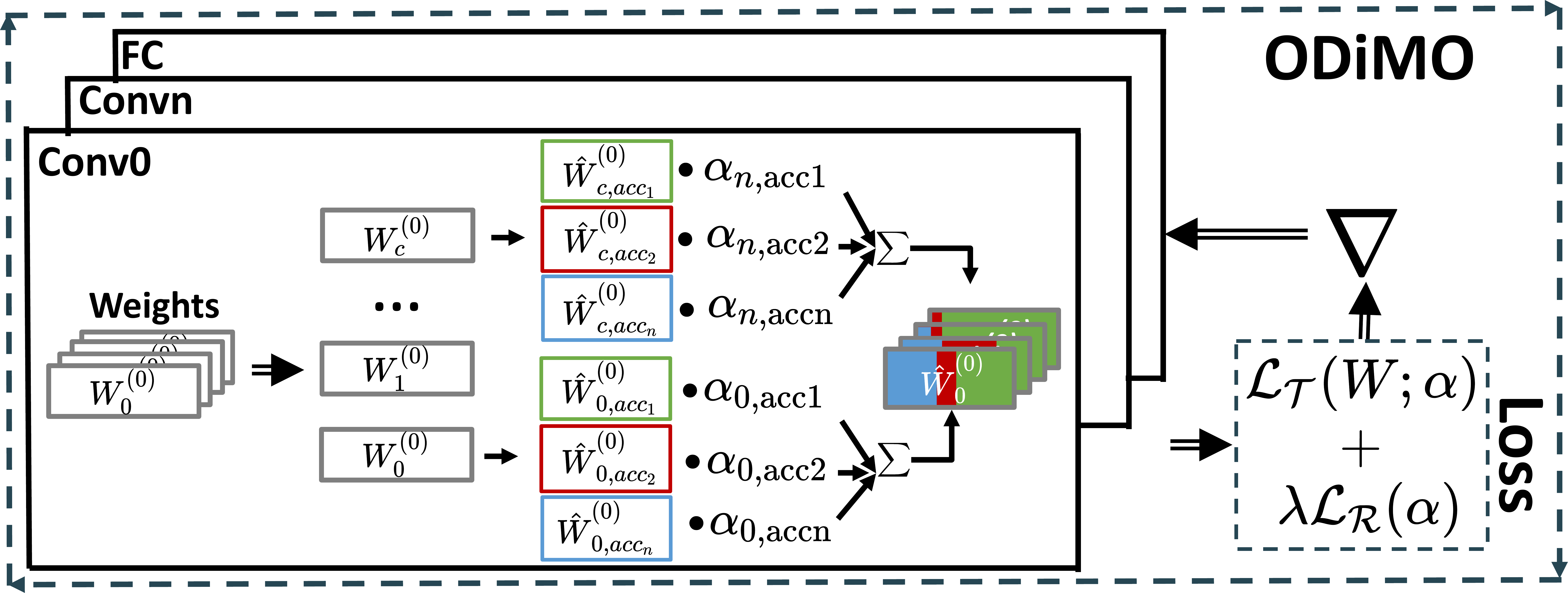}
  \vspace{-0.2cm}
  \caption{DNAS-like optimization scheme at training time.}
  \label{fig:flow}
   \vspace{-0.6cm}
\end{figure}

As shown in Fig.~\ref{fig:flow}, for each layer $l$ of a DNN, the weight tensor $W^{(l)}$ is fake-quantized multiple times, simulating the data format supported by \textit{all available accelerators}. Namely, we generate $\hat{W}^{(l)}_{\text{acc}_i}, \forall i \in [1,N]$ different fake-quantized copies of the weights.
Each of them is paired with a vector of trainable parameters $\alpha^{(l)}_{\text{acc}_i} \in \mathbb{R}^{C^{(l)}_{\text{out}}}$, where $C^{(l)}_{\text{out}}$ is the number of output channels in the $l$-th layer.
For each channel $c$ in $C^{(l)}_{\text{out}}$ we then compute the \textit{effective weights} as:
\begin{equation}\label{eq:eff_w}
    \hat{W}^{(l)}_{c} = \sum_{i=1}^{N} \bar{\alpha}^{(l)}_{c, \text{acc}_i}\hat{W}^{(l)}_{c, \text{acc}_i}
    \vspace{-0.2cm}
\end{equation}
where $\bar{\alpha}^{(l)}_{c, \text{acc}_i} =  \mathrm{sofmax}(\alpha^{(l)}_{c, \text{acc}_i}, \tau)$  and $\tau$ is the softmax temperature.
The aggregated effective weight tensor $\hat{W}^{(l)}$ for layer $l$ is obtained concatenating the $\hat{W}^{(l)}_{c}$ tensors of Eq.~\ref{eq:eff_w} over the output channel dimension.

Using $\hat{W}^{(l)}$ in place of $W^{(l)}$ for all layers, ODiMO solves a continuous relaxation of the multi-accelerator mapping problem. In practice, each layer's output becomes a ``mix'' of what would be produced by all available accelerators, given their quantization formats. The importance of each accelerator in the mix is controlled by $\alpha^{(l)}_{\text{acc}_i}$. 
We can then train $\alpha^{(l)}_{\text{acc}_i}$ as in DNAS, to learn which mapping
provides the best accuracy vs inference cost trade-off for a given channel.

Specifically, the DNN with fake-quantized weights is inserted in a training loop which optimizes:
\begin{equation}\label{eq:dnas}
    \min_{W, \alpha} \mathcal{L_T}(W; \alpha) + \lambda \mathcal{L_R}(\alpha)
    \vspace{-0.3cm}
\end{equation}
where $\mathcal{L_T}$ is the standard task loss, $W = \{W^{(l)}\}, \forall l $ is the set of DNN weights, and $\alpha = \{\alpha^{(l)}_{acc_i}\}, \forall l,i $ is the set of parameters that determine the bit-width assignment for each channel, and consequently its mapping to one of the available accelerators.
Lastly,  $\mathcal{L_R}$ is an additional loss term that models the cost of the DNN execution (e.g., energy) as a function of the mapping decisions, and  $\lambda$ is a scalar regularization strength that controls the balance between the two loss terms. 

We formulate $\mathcal{L_R}$ differently when optimizing for energy or latency. For latency, we minimize:
\begin{equation}\label{eq:reg_loss}
    \mathcal{L}_{\mathcal{R}} = \sum_{l} M^{(l)}\mathrm{,\ }M^{(l)} = \max(LAT^{(l)}_{1},...,LAT^{(l)}_{n})
    \vspace{-0.2cm}
\end{equation}
where $LAT^{(l)}_{i}(\alpha)$ is a differentiable model of the $i$-th accelerator's latency for layer $l$, as a function of the channels assigned to it, detailed in Sec.~\ref{sec:hw_models}. $M^{(l)}$ is the latency of the entire layer, assuming that the accelerators run in parallel, which except for thermal effects, which are generally negligible for low-power SoCs like DIANA, is the optimal choice for both time and energy reduction, as it minimizes idle consumption. In practice, since we need a fully-differentiable loss term, we substitute the $\mathrm{max}$ operation of Eq.~\ref{eq:reg_loss} with its smooth differentiable approximation. For energy reduction, instead, we use the following model:
\begin{equation}\label{eq:reg_loss_en}
    \mathcal{L}_{\mathcal{R}} = \sum_l \sum_i P_{act,i}\cdot LAT^{(l)}_{i} + P_{idle,i}\cdot (M^{(l)} - LAT^{(l)}_{i})
    \vspace{-0.2cm}
\end{equation}
where $P_{act,i}$ and $P_{idle,i}$ are the average active and idle power consumption of the $i$-th accelerator.

At the end of training, we \textit{discretize} the mapping. Namely, for each channel, we select the accelerator corresponding to the largest $\alpha^{(l)}_{c, \text{acc}_i}$.

\begin{figure}[t]
  \centering
  \includegraphics[width=.85\columnwidth]{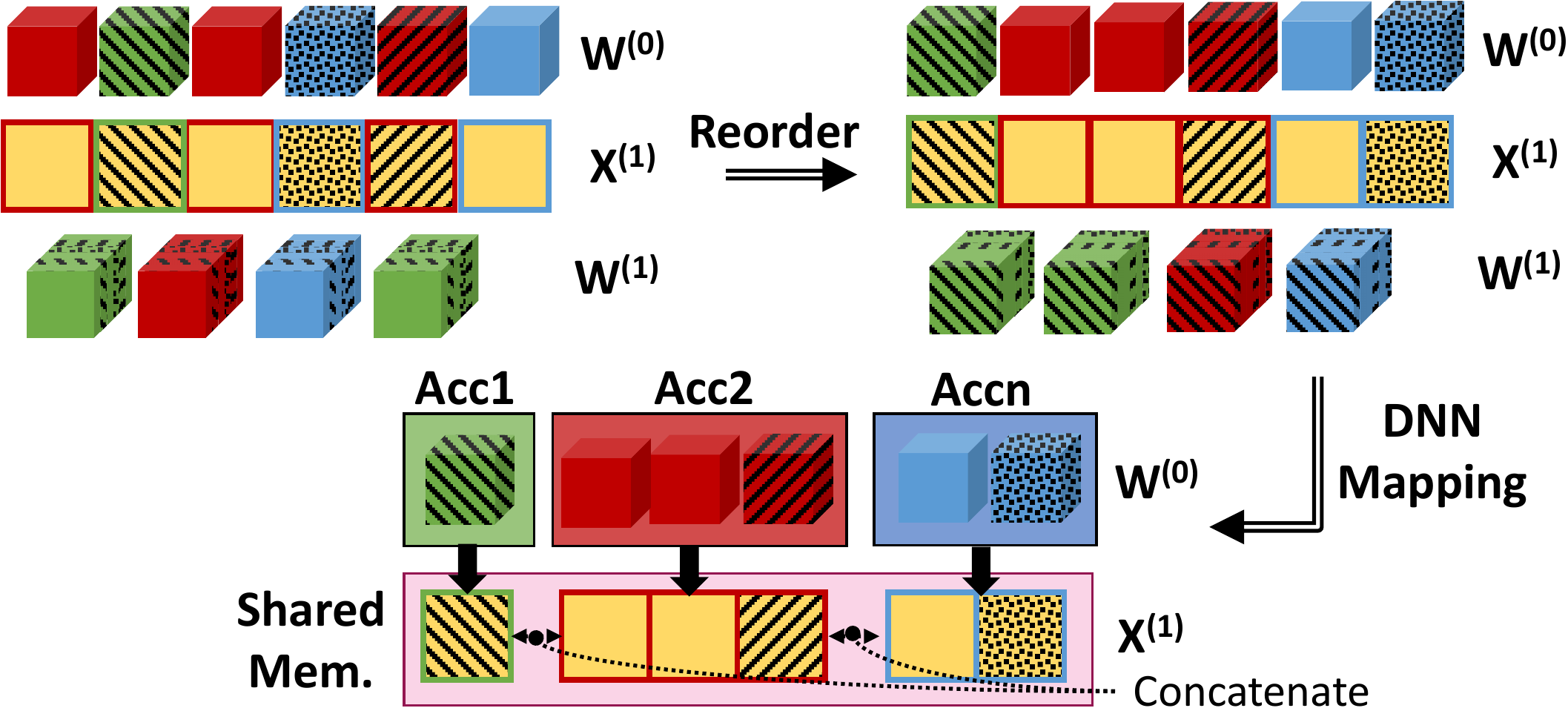}
  \vspace{-0.2cm}
  \caption{Final layer re-organization pass to support partitioning.}
  \label{fig:ch-map}
  \vspace{-0.6cm}
\end{figure}

However, the channels assigned to the same hardware are in general not consecutive, which would complicate the merging of partial outputs. Therefore, a layer transformation pass is applied to the DNN before deployment on the target SoC, shown in Fig.~\ref{fig:ch-map}.
Activation channels are represented as side-by-side squares for better visualization, and colors used for weights filters and for activation outlines indicate the assignment to a given accelerator.
Black patterns are added to some filters/output slices to clarify the process.
The top-left part of the figure shows an example of ODiMO output.
On the top-right, the channels in $X^{(1)}$ and the corresponding filters in $W^{(0)}$ are \textit{reordered}, grouping together those that will be dispatched to the same accelerator, to enable a simple concatenation of outputs.
To preserve the network functionality, the weights of the next layer $W^{(1)}$  are also reordered across the \textit{input} channels dimensions.
After this transformation, the layer is effectively split into $N$ independent sub-layers that can be deployed in parallel onto the $N$ available accelerators, without requiring any data-marshaling overhead to aggregate their outputs (bottom of Fig.~\ref{fig:ch-map}).
\subsection{Training Details} \label{subsec:training}
In this work, we apply ODiMO to the DIANA multi-accelerator SoC of~\cite{ueyoshi2022diana}, presented in Sec.~\ref{sec:soc}.
This section reports the HW-specific details of our implementation. Note that the general approach is orthogonal to most of these details.

Given a pre-trained floating-point DNN, we first fold Batch Normalization (BN) layers with Conv/FC, since the DIANA accelerators do not implement BN in hardware.
Then, we apply fake-quantization following the scheme of~\cite{verhoef2019fq}:
\begin{equation}\label{eq:w_quantizer}
    Q(x) = \frac{e^{s}}{2^{n-1}-1} \cdot \mathrm{round}(2^{n-1} - 1 \cdot \mathrm{clip}(x, -1, 1))
    \vspace{-0.1cm}
\end{equation}
where $s$ is a trainable scale parameter and $n$ is the bit-width.
With $n = 2$, Eq.~\ref{eq:w_quantizer} performs ternarization, i.e., the quantization format of DIANA's AIMC accelerator weights, while we use $n=8$ for the digital accelerator weights. Concerning activations, the AIMC and digital blocks have slightly different formats on 7- and 8-bit respectively. During the optimization phase, we use the worst case of the two (7-bit) as fake-quantization bit-width for layers' inputs/outputs.
As long as the DNN is appropriately fine-tuned (see below), we found this approximation not to degrade our results.

The fake-quantized DNN is optimized with the procedure of Fig.~\ref{fig:flow} until convergence, with an early-stop mechanism.
Then, after discretizing the final channel assignment to each accelerator, the model is fine-tuned based on the task loss term $\mathcal{L}_{\mathcal{T}}$ only. In this phase, we use the exact quantization format also for activations, i.e., shared data are stored on 8-bit but the AIMC accelerator D/A and A/D converters are on 7-bit, effectively truncating the LSB of inputs/outputs.
\subsection{Hardware Models} \label{sec:hw_models}
The differentiable latency models plugged in Eq.~\ref{eq:reg_loss} and \ref{eq:reg_loss_en} are key elements of the proposed method.
Latency modeling has been studied extensively in recent NAS literature. A common approach~\cite{proxylessnas_2018} uses a small NN model trained on many profiled layers to predict latency based on the layer geometry. Although this method is compatible with ODiMO, given the predictability of DIANA's AIMC and Digital accelerators execution, we found that using simpler analytical models that account for the respective parallelism and dataflow
yields good-enough results while making the optimization faster.

These simplified models neglect non-idealities such as memory stalls, tiling overheads for large activation tensors, and programming overheads.
However, comparing them with hardware measurements on a wide set of layer configurations, we verified that they can preserve rank well, i.e., it generally holds that if $LAT_{predicted}^1 < LAT_{predicted}^2$, then $LAT_{hw}^1 < LAT_{hw}^2$,
which makes them usable for mapping decisions.
For the AIMC accelerator, our latency model is:
\begin{align*}
    LAT_{aimc}^{(l)}(\alpha) = &\lceil \frac{C_{in}^{(l)} \times f_x^{(l)} \times f_y^{(l)}}{1152} \rceil \lceil \frac{C_{out}^{(l)}(\alpha)}{512} \rceil \times o_x^{(l)} \times o_y^{(l)}  + \\
                      &2 \times 4 \times C_{in}^{(l)} \times \lceil \frac{C_{out}^{(l)}(\alpha)}{512} \rceil
\end{align*}
\vspace{-0.1cm}
\looseness=-1
where $C_{in}^{(l)}$, $o_x^{(l)}$/$o_y^{(l)}$ and $f_x^{(l)}$/$f_y^{(l)}$ are the layer's input channels, output spatial dimensions and kernel sizes respectively, and the two addends correspond to the cycles of the computation and of the DMA transfer to populate the weights respectively.
Note that this model depends on the optimization choices ($\alpha$) through $C_{out}$, since ODiMO assign a variable number of output channels to the AIMC accelerator.
The digital accelerator model uses the same two terms:
\begin{align*}
    LAT_{dig}^{(n)}(\alpha) = &\lceil \frac{C_{out}^{(l)}(\alpha)}{16} \rceil \lceil \frac{o_y^{(l)}}{16} \rceil \times C_{in}^{(l)} \times o_x^{(l)} \times f_x^{(l)} \times f_y^{(l)} + \\
                    &C_{in}^{(l)} \times C_{out}^{(l)}(\alpha) \times f_x^{(l)} \times f_y^{(l)}
\end{align*}
\vspace{-0.5cm}

Numeric constants in the two models depend on the sizes of the respective processing element arrays. We do not count activation transfers, because we assume that they are always stored in the shared $L_1$ scratchpad memory. 

%% file: text/Results.tex
\begin{figure*}[t]
  \centering
  \includegraphics[width=0.85\textwidth]{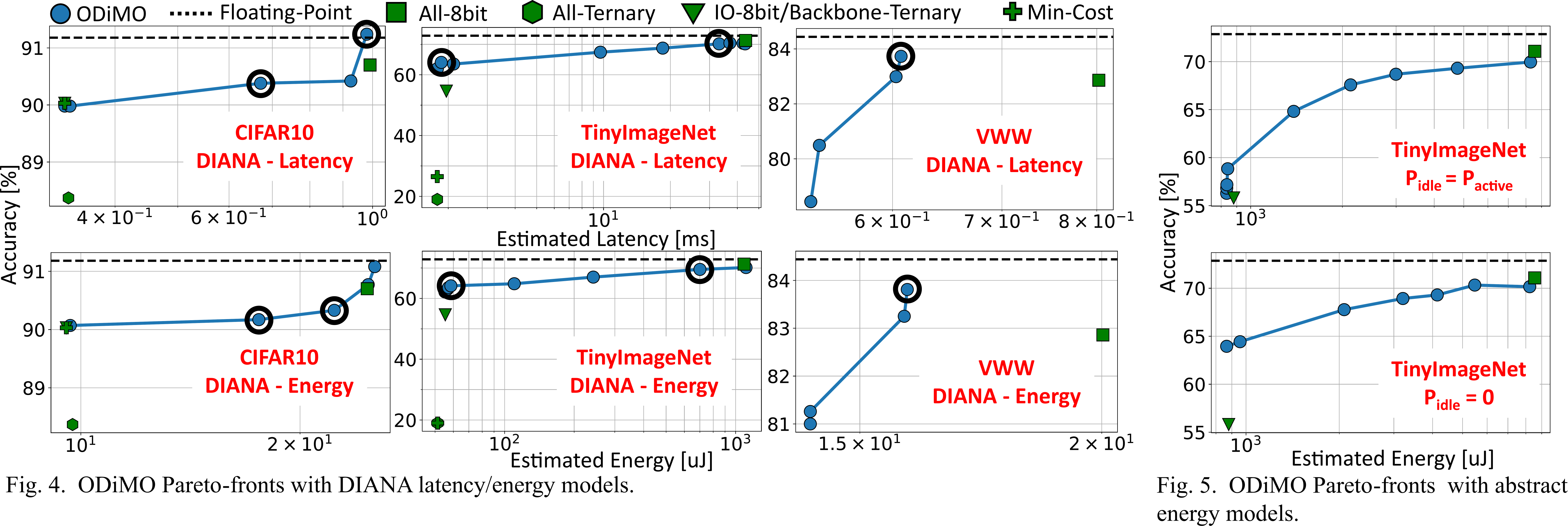}
  \label{fig:fronts}
  \vspace{-.7cm}
\end{figure*}
%
%
%
\section{Experimental Results}
\subsection{Setup}
We benchmark ODiMO on three edge-relevant computer vision tasks and DNNs:
i) image classification on CIFAR-10, with ResNet20~\cite{he2016deep} as reference model; 
ii) image classification on the 200-classes Tiny-ImageNet~\cite{tiny-imagenet}, with ResNet18~\cite{he2016deep}; iii) person detection on Visual Wake Word (VWW), which is based on the MSCOCO 2014 dataset, with a MobileNet-V1 with $0.25\times$ width-multiplier~\cite{mlperf-tiny}.
We pre-train and fine-tune all DNNs using the same epochs and hyper-parameters of the reference papers.
ODiMO is written in Python 3.9 and PyTorch v1.11. 
To deploy our networks on DIANA~\cite{ueyoshi2022diana}, we adapted the open-source DORY~\cite{dory} framework.

We compare ODiMO with several baseline mapping alternatives. Specifically, our baselines are: i) \textit{All-8bit} and \textit{All-Ternary}, the trivial mappings that use only the DIANA digital and AIMC accelerators, respectively; ii) \textit{IO-8bit/Backbone Ternary}, a heuristic solution from~\cite{ueyoshi2022diana} that maps the first/last layers to the 8-bit accelerator, and the rest to the AIMC one, based on the rule-of-thumb that aggressively quantizing layers close to the input and the output of the network often causes large accuracy drops; iii) \textit{Min-Cost} an optimized deterministic mapping that uses the same channel-wise partitioning of ODiMO, with the sole goal of minimizing latency or energy without taking accuracy into account. Namely, it statically maps channels of each layer to the AIMC and digital accelerators, before training, to minimize Eq.\ref{eq:reg_loss} or Eq.\ref{eq:reg_loss_en}. In case of equivalent solutions, digital channels are maximized since this is expected to improve accuracy.

\looseness=-1
For the MobileNetV1 on VWW, we only optimize the mapping of pointwise and standard convolutions (and FC layers), since in DIANA, depthwise convolutions can only be executed on the digital accelerator. Further, all baselines that use the AIMC accelerator are not reported for VWW because their training could not converge, resulting in random predictions.
\subsection{Search-Space Exploration}\label{sec:exploration}
Fig. 4 show the results obtained with ODiMO on the three benchmarks, in the accuracy versus estimated latency (top row) and accuracy versus estimated energy (bottom row) spaces, with latency and energy computed using the DIANA's models described in Sec.~\ref{sec:hw_models}.
Each ODiMO point is obtained repeating the training procedure of Sec.~\ref{subsec:training} with a different regularization strength ($\lambda$) and using either the energy or latency regularizer. We also report the baselines in green and the floating point DNN accuracy as a horizontal dashed line.
In all graphs, \textit{baselines are either dominated or on the Pareto frontier}, demonstrating the effectiveness of our approach. Additionally, ODiMO produces a \textit{rich set of intermediate Pareto-optimal solutions} that could not be obtained otherwise.

With the DIANA cost models, ODiMO can trade-off the estimated latency and accuracy (-$32\%$ latency, -$0.32\%$ accuracy) w.r.t. the All-8bit baseline on CIFAR-10 (3rd blue dot from the right in the top-row figure). Moreover, energy can be reduced by $29\%$ when accepting a $0.53\%$ accuracy drop (4th point from the right in bottom-row).
On TinyImageNet, our tool discovers solutions spanning more than one order of magnitude on the x axis, that can reduce the estimated latency/energy by $15\%/35.6\%$ and $77.8\%/77.7\%$ for a drop of $<$$2\%$ and $<$$5\%$ accuracy w.r.t. the 8bit baseline, respectively. Lastly, on VWW, ODiMO achieves up to $24.3\%/20.8\%$ latency/energy reduction while improving accuracy by $0.87\%/0.95\%$ w.r.t All-8bit.

Fig. 5 shows the independence of ODiMO from the DIANA SoC specifics.
For the sake of space, the figure shows results only on Tiny-ImageNet, and demonstrates how ODiMO is able to find a rich collection of Pareto optimal mappings even with different hardware cost models.
Results are obtained considering two abstract models, not related to any specific HW, retaining from DIANA only the presence of two accelerators working with ternary and 8-bit data precision respectively.
These models assume that the latency of both accelerators is simply \textit{proportional to the number of operations}, and that the active power of the 8-bit accelerator is 10 times higher than the ternary one ($P_{act,8} = 10\cdot P_{act,ter}$).
Then, for the first model, we assume $P_{idle} = P_{act}$ for both accelerators (no shutdown), while, for the second, we consider $P_{idle} = 0$ (ideal shutdown).

The top graph of Fig. 5 shows ODiMO mappings obtained with the first model.
Note that in this corner case, energy and latency minimization coincide, since substituting $P_{idle} = P_{act}$ for all accelerators in Eq.~\ref{eq:reg_loss_en} yields Eq.~\ref{eq:reg_loss}, except for a constant.
The bottom of Fig. 5, instead, shows the results obtained with $P_{idle,i} = 0$.
The two graphs only show accuracies $> 55\%$; the all-ternary and min-cost baselines are not shown, as they reach too low accuracy (see middle graphs in Fig. 4).
With these two models, ODiMO reduces energy respectively by $44.2\%/51.5\%$ for a drop of $<$$2\%$ accuracy w.r.t. the 8bit baseline.
\subsection{DIANA Deployment}
This section analyzes the results of deploying a subset of the solutions from Fig. 4 on the DIANA SoC, running at a frequency of 260 MHz, substituting modeled with measured latency and energy.
For each benchmark, we deploy the All-8bit and Min-Cost baselines and 
a selection of ODiMO results (highlighted with a black circle in Fig. 4). We select two points from the latency Pareto-front (Large-Lat and Small-Lat) and two from the Energy one (Large-En, Small-En) for all benchmarks except VWW, where given the smaller search space, we deploy a single point from each graph.
For all DNNs, we report in Table \ref{tab:networkperformance} accuracy, latency, energy consumption, the percentage of time each accelerator is utilized during an end-to-end inference (\textit{D./A. util.}), and the percentage of channels executed on the AIMC accelerator, i.e., the fraction $\nicefrac{C_{out}^{aimc}}{C_{out}}$ for the whole network (\textit{A. Ch.}).
\input{tables/Table_Networks}
\renewcommand{\thefigure}{6}
\begin{figure}[t]
  \centering
  \includegraphics[width=0.85\columnwidth]{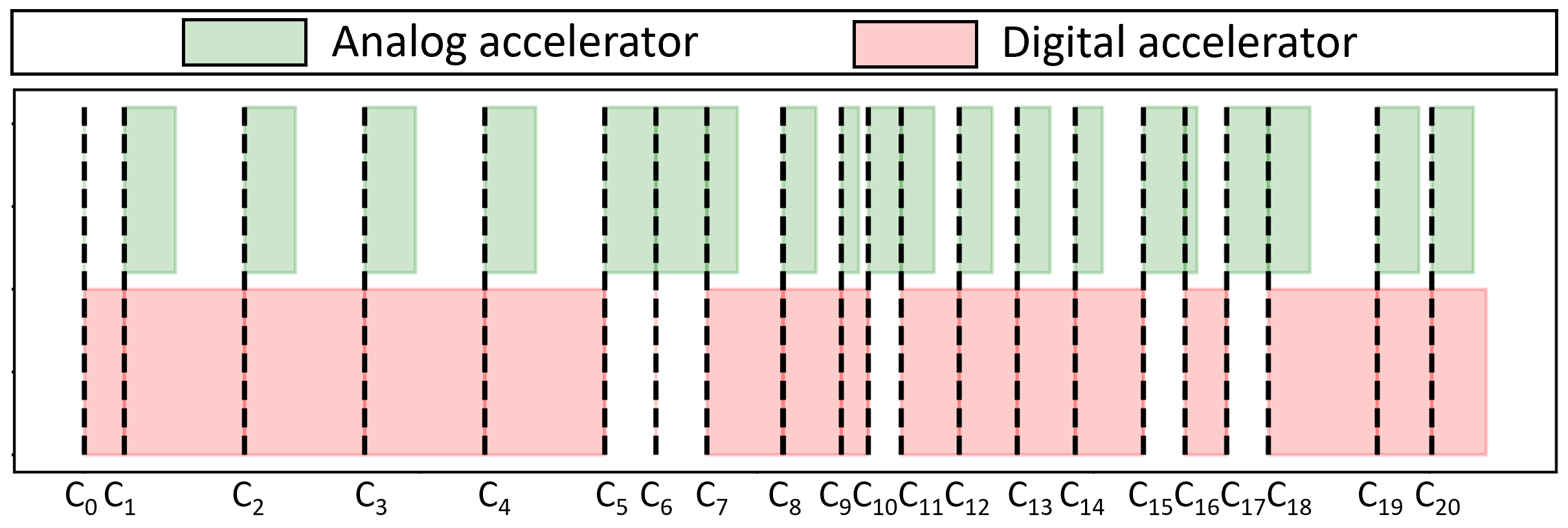}
  \vspace{-0.3cm}
  \caption{Utilization of accelerators on convolutional layers of ODiMO-Small-En on CIFAR-10. ($C_i = $ $i$-th Conv. layer).}
  \label{fig:utilization}
  \vspace{-0.7cm}
\end{figure}
On CIFAR-10, ODiMO-Small-En reduces energy by $33\%$ w.r.t All-8bit, for a limited accuracy drop (-$0.53\%$). This result, which is compatible with the $29\%$ reduction estimated by the model (see Sec.~\ref{sec:exploration}), is achieved offloading a large portion of the channels (72.9\% of the total) to the analog accelerator.
Further, the digital and AIMC accelerators are active for $76.2\%$ and $60\%$ of the inference time. Fig.~\ref{fig:utilization} shows a breakdown of the utilization of both accelerators throughout an inference with this DNN. For almost the $40\%$ of the time, both accelerators work simultaneously, demonstrating that splitting layers between the two is beneficial to reduce energy consumption, while keeping an almost constant accuracy.

On TinyImageNet, ODiMO-Large-En suffers an accuracy drop compared to All-8bit (-$1.75\%$), but improves the energy by 1.43$\times$, while ODiMO-Small-En achieves $37.63\%$ higher accuracy compared to Min-Cost, at the cost of only 1.12$\times$ higher energy consumption.
Both solutions exploit the analog accelerator for a large portions of the DNN channels, $34.2\%$ and $96.5\%$, respectively.
It is also worth mentioning that Min-Cost, which offloads only an additional $1.5\%$ of the network to the AIMC accelerator compared to ODiMO-Small-En, fails in reaching a good accuracy; this is because the Min-Cost mapping is built without taking into account accuracy, contrary to our method.
Further, notice that ODiMO-Large-Lat effectively improves latency compared to All-8bit (1.27$\times$ faster) but at same time fails in reducing the energy consumption (1.27$\times$ less efficient) demonstrating the need to optimize energy and latency with two different tailored models, depending on the specific design goals.

On VWW, despite lower benefits, ODiMO-En shows a higher accuracy compared to the All-8bit (+$0.95\%$) solution with $7\%$ lower energy consumption, Pareto dominating it.

%% file: tables/Table_Networks.tex
\begin{table}[t]
\caption{Deployment on Diana of selected solutions from Fig. 4}
\vspace{-0.2cm}
\label{tab:networkperformance}
\resizebox{\columnwidth}{!}{
\begin{tabular}{l|l|l|l|l|l|l}
                           & Network      & Acc. & lat. {[}ms{]}& E. {[}uJ{]} & D./A.   util. & A. Ch. \\ \hline
\multirow{6}{*}{Cifar10} & All-8bit & 90.70         & 1.55 & 38.71     & 100\% / 0\%         & 0\%     \\
& ODiMO Large - Lat   & 91.24         & 1.55 & 43.20          & 100\% / 21.0\%      & 5.6\% \\
& ODiMO Small - Lat   & 90.38         & 1.07  & 34.43        & 100\% / 44.8\%     & 51.8\% \\
& ODiMO Large - En   & 90.33         & 1.05 & 33.43          & 100\% / 43.1\%      & 50.3\% \\
& ODiMO Small - En   & 90.17         & 0.80 & 25.94          & 76.2\% / 60.0\%     & 72.9\% \\
& Min Cost  & 90.06         & 0.47 & 13.57         & 9.5\% / 93.6\%      & 97.5\%  \\ \hline
\multirow{6}{*}{TinyI.}   & All-8bit & 71.29         & 94.44 & 2357.3        & 100\% / 0\%         & 0\%     \\
& ODiMO Large - Lat   & 70.16         & 73.92 & 2999.8         & 100\% / 8.2\%         & 23.8\% \\
& ODiMO Small - Lat    & 64.07         & 4.32 & 139.2        & 25\% / 87.8\%       & 99.0\% \\
& ODiMO Large - En   & 69.54         & 63.55 & 1648.18     & 100\% / 9.4\%      & 34.2\% \\
& ODiMO Small - En   & 64.14         & 5.05 & 141.25        & 20\% / 84.4\%     & 96.5\% \\
& Min Cost  & 26.51         & 4.07 & 125.96          & 30\% / 89.7\%       & 98\%      \\ \hline
\multirow{3}{*}{VWW}       & All-8bit & 82.86         & 3.05 & 76.18          & 100\% / 0\%         & 0\%     \\
& ODiMO - Lat        & 83.73         & 2.80 & 71.29          & 100\% / 17.8\%     & 39.1\% \\
& ODiMO - En   & 83.81         & 2.79 & 70.74          & 100\% / 17.71\%     & 40\% \\
                           \hline
\end{tabular}
}
\vspace{-0.3cm}
\end{table}

%% file: text/Conclusion.tex
\section{Conclusions}
We have introduced ODiMO, a tool that partitions a DNN execution at fine grain among multiple accelerators with incompatible quantization formats. To do so, it formulates the problem as a mixed-precision bit-width assignment and uses a DNAS-like approach to optimize the mapping while training the DNN weights. With results on different benchmarks and DNN architectures, we have shown that ODiMO can obtain rich Pareto-fronts in both the accuracy vs energy or latency spaces, and reduce energy by up to $33\%$ with limited accuracy drops compared to a single-accelerator solution. Future work will concentrate on building more accurate hardware models and supporting also activations quantization requiring format conversions whose cost need to be modeled.